\documentclass[runningheads]{llncs}
\usepackage[T1]{fontenc}
\usepackage{graphicx}
\usepackage{listings}
\usepackage{adjustbox}
\usepackage[margin=1.5in]{geometry}    
\usepackage[english]{babel}
\usepackage[utf8]{inputenc}
\usepackage{algorithm}
\usepackage{arevmath}     
\usepackage[noend]{algpseudocode}
\usepackage{multirow}
\usepackage{graphicx}
\usepackage{subcaption}
\usepackage{capt-of}
\usepackage{makecell}
\usepackage{xcolor}
\usepackage{amssymb}
\usepackage{pifont}
\usepackage[misc]{ifsym}

\newcommand{\cmark}{\ding{51}}%
\newcommand{\xmark}{\ding{55}}%
\usepackage{perpage} 
\MakePerPage{footnote} 

\makeatletter
\newcommand*\NOINDENT{\@@par   
      \@totalleftmargin\z@ \@listdepth\z@ \rightmargin\z@
}
\makeatother

\begin{document}
\title{ZeFaV: Boosting Large Language Models for Zero-shot Fact Verification}

\author{Son T. Luu\inst{1,2,\thanks{Equal contribution.}} \and 
Hiep Nguyen\inst{1,\footnotemark[1]} \and Trung Vo\inst{1} \and Le-Minh Nguyen\inst{1,\thanks{Corresponding author.}}}
\institute{Japan Advanced Institute of Science and Technology, Ishikawa, Japan \and VNU University of Information Technology, Ho Chi Minh City, Vietnam \\
\email{\{sonlt,hiepnkv,trungvo,nguyenml\}@jaist.ac.jp}}

\maketitle     

\begin{abstract}
In this paper, we propose ZeFaV - a zero-shot based fact-checking verification framework to enhance the performance on fact verification task of large language models by leveraging the in-context learning ability of large language models to extract the relations among the entities within a claim, re-organized the information from the evidence in a relationally logical form, and combine the above information with the original evidence to generate the context from which our fact-checking model provide verdicts for the input claims. We conducted empirical experiments to evaluate our approach on two multi-hop fact-checking datasets including HoVer and FEVEROUS, and achieved potential results results comparable to other state-of-the-art fact verification task methods.

\end{abstract}

\section{Introduction}
\label{intro}
\vspace{-1em}
Fact-checking involves determining the truth of a claim \cite{guo2022survey}. With the rise of the Internet, fake news and misinformation have become significant issues on social media \cite{nakov2021automated}. Fact-checking is time-consuming as it requires detecting the claim, gathering evidence from verified sources, and comparing information to assess its accuracy. According to \cite{nakov2021automated}, the application of AI to automatically fact-verification allows organizations to perform fact-checking faster and more comprehensively. With the given information as the claim and the relevant evidence, the system must determine the veracity of the claim based on the understanding of the claim context and the evidence related to the claim. Nonetheless, there are four main challenges for the automated fact verification task. First, the claim is ambiguous when there are several ways to interpret the meaning of the claim \cite{nakov2021automated}. Second, the data artifacts and bias in the annotated dataset can lead to bias in system predictions \cite{guo2022survey}. Third, the current state-of-the-art system is limited to contextual information such as external knowledge sources and users' created information \cite{guo2022survey}. Last, the making of annotated data for fact-checking is costly, time-consuming, and potentially biased (according to \cite{pan-etal-2023-fact}). Since the limitation in the scale of training data, the utilization of a pre-trained language model deals with domain-sensitive problems, in which the pre-trained language model on a specific domain cannot cover other domains without re-training the model \cite{cao2023large}. 

To address these challenges, applying Large Language Models (LLMs) in fact verification tasks is promising due to their strong comprehension of human language \cite{cao2023large}. Previous works like ProgramFC \cite{pan-etal-2023-fact}, QAChecker \cite{pan-etal-2023-qacheck}, and InfoRE \cite{cheng2024information} utilize LLMs for fact verification and show potential results. However, LLM outputs are not always consistent \cite{cao2023large}, and the reasoning process needs support for the chain of thought (CoT) \cite{wei2022chain}. Additionally, hosting and executing LLMs is costly and inefficient \cite{minaee2024large}. The authors in \cite{pan-etal-2023-qacheck} recommend using locally run and open-source LLMs like LLaMa \cite{touvron2023llama} instead of external API-based LLMs like InstructGPT or GPT-4 \cite{achiam2023gpt}. Therefore, our work aims to enhance LLMs for fact-checking by enriching and guiding their understanding and reasoning processes. We propose ZeFaV, a framework that utilizes few-shot relation extraction and reorganized information (InfoRE) to boost the performance of LLMs for the fact verification task. Few-shot Relation extraction identifies entities in the claim and evidence and their relationships, using these relations to guide LLMs in understanding and verifying the claim while InfoRE reconstructs the evidence provided with the claim to help LLMs better comprehend it. Our empirical results show that our proposed method improves the ability of LLMs for the fact verification task. All of our experiments are based on zero-shot learning on LLMs.
\vspace{-1em}

\section{Related Works}
\label{relatedwork}
\vspace{-1em}
The in-context learning ability of large language models demonstrated the effectiveness in verifying the truthfulness of complex claims. ProgramFC \cite{pan-etal-2023-fact} recently used the in-context learning ability of large language models to break down the input claims into reasoning sub-tasks and then obtain the verdicts by tackling each reasoning sub-task. Using few-shot learning, it decomposes claims into Python-like programs with functions to question, verify, and predict facts. ProgramFC demonstrates that decomposing claims into smaller tasks is more effective than a one-step prediction approach, reducing the cognitive load on the language model and enhancing its fact-checking capability. Another approach, the QACheck system \cite{pan-etal-2023-qacheck}, includes five components, each component is a LLM with a specific task for the verification process. These LLMs leverage their ability to generate text and learn from context to ask and answer key questions that determine the claim's truthfulness. In addition, InfoRE \cite{cheng2024information} improves these models' reasoning abilities by restructuring evidence into MindMap forms, which effectively represent knowledge and concepts \cite{buzan2010mind}. This reorganized evidence used alone or with original data, improves claim verification. Experiments show that InfoRE achieves comparable results in fact-checking, demonstrating the effectiveness of this information reorganization method. Finally, a recent study from \cite{li2023rethinking} shows that large language models excel in few-shot relation extraction (FSRE) by producing linearized strings that encode entity pairs and their relations. 
\vspace{-1em}
\section{Proposed Methodology}
\label{method}
\vspace{-1em}

Giving a sample \textit{(c, E)}, where \textit{c} is a claim sentence, and \textit{E} is a set of evidence relevant to \textit{c}. The fact verification task aims to obtain the result \textit{v} such that \textit{v} $\in$ \{\textit{True, False}\}. The \textit{True} indicates the claim is correct (supports), and \textit{False} indicates the claim is not correct (refutes). To solve the task, we proposed the ZeFaV\footnote{The source code is available at: \url{https://github.com/sonlam1102/zefav}} - our zero-shot prompting technique leverages the text re-organizing and relation extraction to enhance the reasoning ability of the LLMs for the Fact verification task. The ZeFaV consists of three main stages, as described below. 

First, we fine-tune the LLMs using the prompt below for the relation extraction task based on the FewRel dataset \cite{gao2019fewrel}. Then, we use this prompt to extract the relation for an input sentence (we leave the \textbf{\textit{\#\#\# Response: }} as blank thus the LLM can generate the corresponding results). 
\lstset{
    basicstyle=\small\ttfamily,
    columns=fullflexible,
    breaklines=true,
    xleftmargin=\dimexpr\fboxsep-\fboxrule,
    xrightmargin=\dimexpr\fboxsep-\fboxrule,
    breakautoindent=false
}
\begin{lstlisting}
### Instruction: Given a sentence, please identify the head and tail entities in the sentence and classify the relation type into one of the appropriate categories; The collection of categories is: [<list_relations_in_FewRel>]; 
Sentence: [CLAIM_SENTENCE] 
### Response: (claim_head, claim_relation, claim_tail)
\end{lstlisting}

Second, we referred to the closure definition, which is a function that combines the bundled with their surrounding states, to find the evidence relations that are related to the claim and remove the redundant relations in the evidence, as shown in Algorithm \ref{alog_closure_extraction}. 
\vspace{-2em} 

\begin{algorithm}[H]
    \caption{Finding a closure of evidence relation from the claim relation}
    \label{alog_closure_extraction}
    \scriptsize
    \begin{algorithmic}
        \Procedure{FindingEvidenceRelations}{Claim\_Rels, Evidence\_Rels}
        \State Evidence\_Rels\_News $\gets$ \{\}
        \State Hypos $\gets$ \{\}
        \State is\_found $\gets$ True 
        \State Hypos $\gets$ Hypos $\cup$ head(claim\_rel) $\cup$ tail(claim\_rel) | claim\_rel $\in$ Claim\_Rels
        \While{is\_found is True}
            \For{evidence\_rel $\in$ Evidence\_Rels} 
                \If{head(evidence\_rel) $\in$ Hypos} 
                    \State Evidence\_Rels\_News $\gets$ Evidence\_Rels\_News $\cup$ evidence\_rel
                    \State Hypos $\gets$ Hypos $\cup$ tail(evidence\_rel)
                \EndIf
                \If{Cannot expand the \textit{Hypos} anymore}
                    \State is\_found $\gets$ False 
                \EndIf
            \EndFor
        \EndWhile
        \State Return Claim\_Rels, Evidence\_Rels\_News
        \EndProcedure
    \end{algorithmic}
\end{algorithm}
\vspace{-2em} 

Third, we apply the InfoRE \cite{cheng2024information} to structure evidence into a more informative textual format, enhancing the in-context learning of these models. From evidence set \textit{E} related to claim \textit{c}, we derive set \textit{E'} containing re-organized evidence, structured using few-shot learning techniques by these models. The prompt template used is as follows:

\lstset{
    basicstyle=\small\ttfamily,
    columns=fullflexible,
    breaklines=true,
    xleftmargin=\dimexpr\fboxsep-\fboxrule,
    breakautoindent=false,
    xrightmargin=\dimexpr\fboxsep-\fboxrule
}

\begin{lstlisting}
Transform the following text into a hierarchical structure that organizes the information in the text into levels. The same level can reflect parallel relationships and indented levels reflect causal relationships. Here are some examples:
The evidence: [EXAMPLE EVIDENCE 1]
The hierarchical structure: [EXAMPLE RE-ORGANIZED EVIDENCE 1]
....
### The evidence: [EXAMPLE EVIDENCE 2]
The hierarchical structure:
\end{lstlisting}

\begin{figure}[H]
\centering
  \includegraphics[width=.9\textwidth]{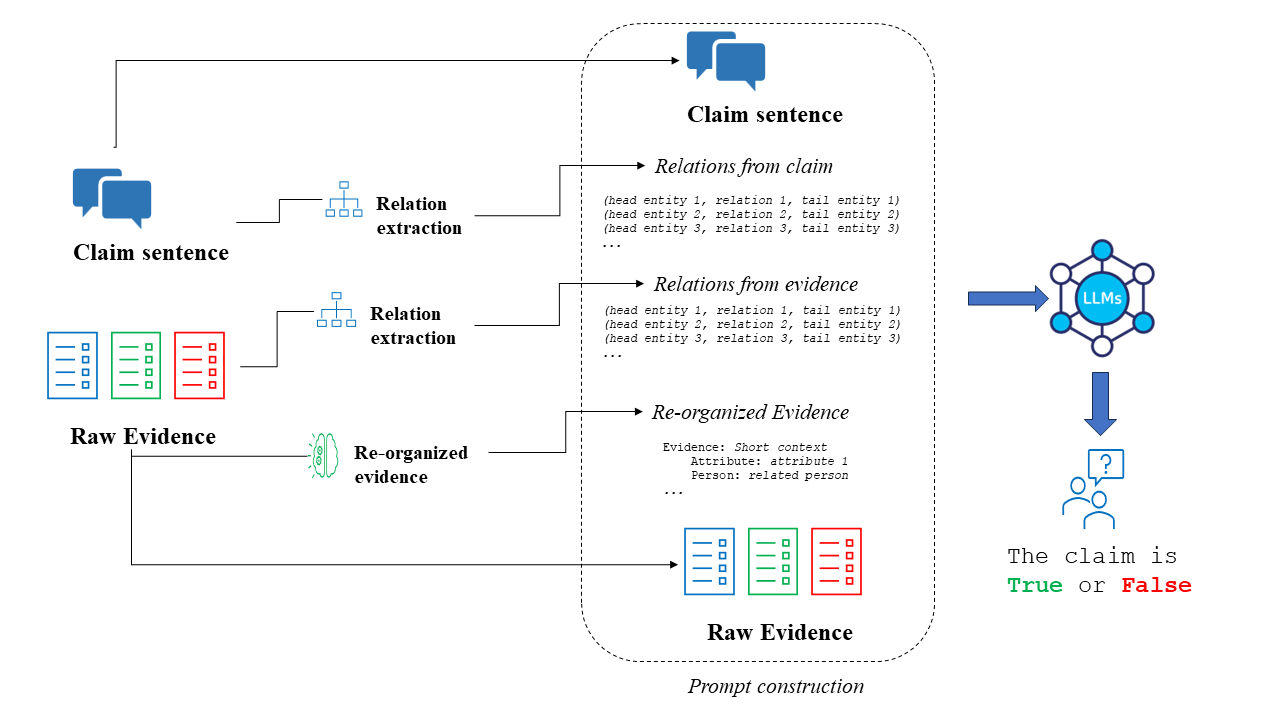}
    \caption{Overview of the ZeFaV framework}
    \label{fig_zerfav_framework}
\end{figure}

Finally, we combined the extracted relations, the reorganized context, and the claim as the prompt below to perform zero-shot Fact verification on LLMs (* representing claim relations, ** representing evidence relations). Figure \ref{fig_zerfav_framework} illustrates the overview framework of ZeFaV. 
\lstset{
    basicstyle=\small\ttfamily,
    columns=fullflexible,
    breaklines=true,
    xleftmargin=\dimexpr\fboxsep-\fboxrule,
    breakautoindent=false,
    xrightmargin=\dimexpr\fboxsep-\fboxrule
}
\begin{lstlisting}
Documents: [The InforRE content]
Context: [The full context evidence in textual form]
Question: [The claim]?
Please answer the question based on Documents, Context, and the following relations. The answer must belong to one of two values: True or False.
    1. The question mentioned the relation between <head entity> and <tail entity> as <type of relation> *
    2. ... 
    3. <head> and <tail> has relation with <type of relation> **
    4. ... 
    Let's think step-by-step.
###The answer is:  
\end{lstlisting}
\vspace{-2em}


\vspace{-3em}
\section{Empirical Results}
\label{results}
\vspace{-0.5em}
\subsection{Main experimental results}
\vspace{-0.5em}
We perform ZeFaV on \textit{Meta-Llama-3-70B-Instruct\footnote{\url{https://huggingface.co/meta-llama/Meta-Llama-3-70B-Instruct}}} and evaluate the performance on the division of HoVer \cite{jiang2020hover} and FEVEROUS-S \cite{aly-etal-2021-fact} datasets by Pan et al. \cite{pan-etal-2023-fact} through the F1-score metric. We run the LLMs with LoRA technique \cite{hu2021lora} quantized to 4 bits, and the maximum length for the generation model is 2,048. Our ZeFaV can run on one NVIDIA A40 GPU with 49GB of memory. For the ProgramFC \cite{pan-etal-2023-fact}, we run the proposed methodology with N=1 on \textit{Meta-Llama-3-70B-Instruct} for comparison with our ZeFaV. Table \ref{tbl_empricial_results} describes our empirical results with ZeFaV compared to other robust methods.
\vspace{-2em}
\begin{table}[H]
    \centering
    \caption{Emprical results of ZeFaV on the HoVer and FEVEROUS dataset}
    \label{tbl_empricial_results}
    \resizebox{.9\textwidth}{!}{
    \begin{tabular}{l|r|r|r|r}
        & \textbf{HoVer (2-hop)} & \textbf{HoVer (3-hop)} & \textbf{HoVer (4-hop)} & \textbf{FEVEROUS-S }\\
        \hline
        Number of claims & 1,126 & 1,835 & 1,039 & 2,962 \\
        \hline
        \textbf{ZeFaV} (w context) & \textbf{77.85} & \textbf{70.61} & \textbf{67.47} & \textbf{86.74} \\
        \textbf{ZeFaV} (w/o context) & 71.31 & 67.41 & 61.62 & 75.97 \\
        \hline 
         ProgramFC \cite{pan-etal-2023-fact} & 76.53	& 70.84	& 66.88	& 86.34 \\
        QACheck \cite{pan-etal-2023-qacheck} \textit{(InstructGPT)} & 55.67 & 54.67 & 52.35 & 59.47 \\ 
        InfoRE \cite{cheng2024information} \textit{(llama2-70B)} & 52.40 & 51.21 & 50.07 & 67.96 \\
        \hline
    \end{tabular}
    }
\end{table}
\vspace{-2em}

According to Table \ref{tbl_empricial_results}, ZeFaV obtained better results than other methodologies on the HoVer dataset. On the FEVEROUS-S dataset, ZeFaV achieved 86.54\% by F1-score, which is better than ProgramFC \cite{pan-etal-2023-fact}, InfoRE \cite{cheng2024information} and QACheck \cite{pan-etal-2023-qacheck}. Since ProgramFC is a few-shot learning method for about 20 examples \cite{pan-etal-2023-fact}, ZeFaV performs efficiently on the HoVer dataset. In comparison with InfoRE, ZeFaV outperforms InfoRE with the LLama architecture on both HoVer and FEVEROUS-S datasets. Besides, the context evidence plays a vital role in the ZeFaV when it increases the performance of the Fact verification task on both HoVer and FEVEROUS-S. It can be seen that ZeFaV showed efficient performance for the Fact verification task with zero-shot learning since it outperforms other methodologies on the same LLama architecture. 

\subsection{Results Analysis}
\vspace{-0.5em}
As shown in Table \ref{tbl_ablation_study}, it can be seen that both InfoRE and Relation help increase the accuracy of LLMs on 2-hop and 4-hop of HoVer. Specifically, on the 3-hops of HoVer, InfoRE helps increase the accuracy while it is slightly decreased when combined with the relation. This is similar to the FEVEROUS-S where the performance slightly decreases when combining InfoRE with relation. In general, InfoRE helps the LLMs in reasoning and understanding the information by re-organizing the data in a compact and concise form. For the relationship, it helps increase the ability of LLMs when combined with InfoRE. However, the performance of ZeFaV when there is only a relation is not as good as InfoRE. This is more clear when there is a lack of evidence context. The accuracy of ZeFaV with InfoRE is better than ZeFaV with relation only. Overall, both InfoRE and Relation help increase the performance of LLMs, and the evidence context also plays a vital role in the zero-shot Fact verification task.
\vspace{-3em}

\begin{table}[H]
    \centering
    \caption{Ablation study on the performance of LLMs for Fact verification}
    \label{tbl_ablation_study}
    \resizebox{.9\textwidth}{!}{
        \begin{tabular}{lcc|rrrr}
        \hline
        & \multicolumn{1}{l}{Relation} & \multicolumn{1}{l|}{InfoRE} & \multicolumn{1}{l}{\textbf{HoVer (2-hop)}} & \multicolumn{1}{l}{\textbf{HoVer (3-hop)}} & \multicolumn{1}{l}{\textbf{HoVer (4-hop)}} & \multicolumn{1}{l}{\textbf{FEVEROUS-S}} \\
    \hline
    \multirow{3}{*}{Has Evidence context} & \cmark                            & \cmark                          & 77.85                             & 70.61                             & 67.47                             & 86.74                          \\
                                          & \xmark                            & \cmark                          & 76.78                             & 72.03                             & 67.64                             & 86.89                          \\
                                          & \cmark                            & \xmark                          & 76.16                             & 69.72                             & 66.70                             & 85.11                          \\
    \hline
    \multirow{3}{*}{No Evidence context}  & \cmark                            & \cmark                          & 71.31                             & 67.41                             & 61.62                             & 75.97                          \\
                                          & \xmark                            & \cmark                          & 71.57                             & 68.30                             & 63.59                             & 76.11                          \\
                                          & \cmark                            & \xmark                          & 61.61                             & 56.64                             & 55.16                             & 59.80      \\
    \hline
    \end{tabular}
    }
\end{table}
\vspace{-4em}

\begin{figure}[H] 
  \begin{subfigure}[b]{0.33\linewidth}
    \centering
    \includegraphics[width=\linewidth]{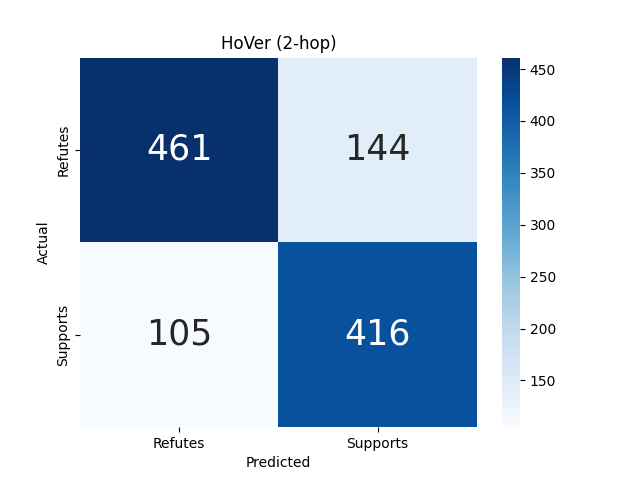} 
  \end{subfigure}
  \begin{subfigure}[b]{0.33\linewidth}
    \centering
    \includegraphics[width=\linewidth]{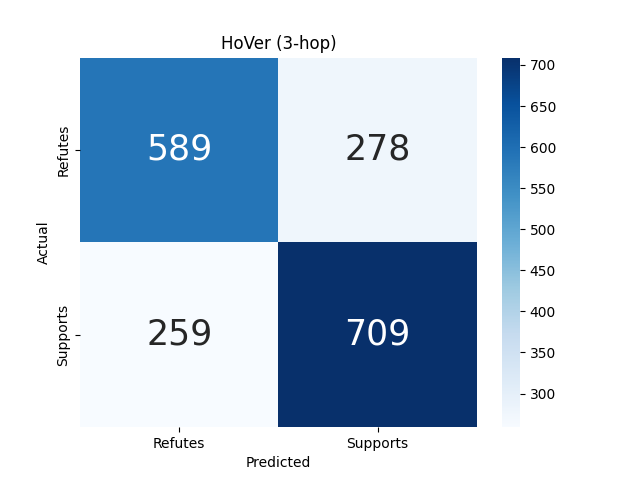} 
  \end{subfigure} 
  \begin{subfigure}[b]{0.33\linewidth}
    \centering
    \includegraphics[width=\linewidth]{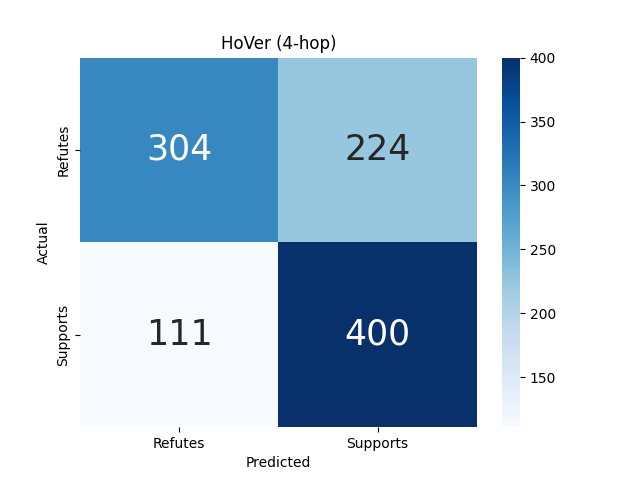} 
  \end{subfigure}
  
  \caption{Confusion matrix of ZeFaV on HoVer}
  \label{fig_confusion_matrix_hover} 
\end{figure}
\vspace{-4em}


\begin{table}[H]
    \centering
    \caption{Performance of ZeFaV on each type of challenge on the FEVEROUS-S dataset}
    \label{tbl_challenge_feverous}
    \resizebox{.9\textwidth}{!}{
    
    \begin{tabular}{l|r|r|r|r|r|r}
    \hline
    \textbf{Type of challenge} & \multicolumn{1}{l|}{\makecell[l]{Search terms \\ not in claim}} & \multicolumn{1}{l|}{\makecell[l]{Multi-hop \\ Reasoning}} & \multicolumn{1}{l|}{\makecell[l]{Combining Tables \\ and Tex}} & \multicolumn{1}{l|}{\makecell[l]{Entity \\ Disambiguation}} & \multicolumn{1}{l|}{\makecell[l]{Numerical \\ Reasoning}} & \multicolumn{1}{l}{Other} \\
    \hline
    \textbf{Number of claims}  & 46                                            & 459                                     & 7                                            & 112                                       & 103                                     & 2,235                     \\
    \hline
    \textbf{F1-score (\%)}     & 82.47                                         & 82.72                                   & 100.00                                       & 82.80                                     & 72.83                                   & 86.77                    \\
    \hline 
    \end{tabular}
    }
\end{table}
\vspace{-2em}
Additionally, according to Figure \ref{fig_confusion_matrix_hover}, the model tends to predict the "refutes" claim to be the "support" claim rather than predicting the "supported" claim to "refutes" claim. The challenge remains in the claim that 4-hops in the HoVer dataset, where the number of wrong predictions in which refuted claims are predicted as supported claims is significant. Besides, it can be seen from Table \ref{tbl_challenge_feverous} that the ZeFaV performs well on almost all types of challenges. For the Numerical Reasoning, ZeFaV is still inefficient where the accuracy by F1-score is 72.83\%.

\vspace{-1em}
\section{Conclusion}
\label{conclusion}
\vspace{-1em}
In this paper, we proposed ZeFaV - a method to address the challenge of improving LLM performance in fact verification using zero-shot learning. Our method leverages relation extraction and InfoRE to enhance LLM fact verification through zero-shot learning with improved evidence representation for CoT learning. The empirical study shows that the ZeFaV performance is efficient when evaluated on the HoVer and FEVEROUS-S datasets. However, from the empirical results, we found that the challenge for this task is the reasoning ability on multiple and complex information such as non-textual and numerical context. Therefore, our future work focuses on improving the ability to reason in complex contexts such as numeric, table, and temporal to increase the accuracy of the Fact verification task.

\bibliographystyle{splncs04}
\bibliography{references}

\end{document}